\documentclass[journal]{IEEEtran}
\usepackage{amsmath}
\usepackage{etoolbox}
\usepackage{adjustbox}
\usepackage{algorithmic}
\usepackage{float}
\usepackage{graphicx}
\usepackage{textcomp}
\usepackage{csquotes}
\usepackage{xcolor,soul}
\usepackage{adjustbox}
\usepackage{hyperref}
\usepackage{multirow}
\usepackage{titlesec}
\usepackage{subcaption}
\usepackage{makecell}
\usepackage{array}
\usepackage{booktabs} 
\usepackage{graphicx} 
\usepackage{amssymb}  
\usepackage{bm}       

\usepackage{caption}
\definecolor{columbiablue}{rgb}{0.61, 0.87, 1.0}
\titlespacing{\subsubsection}{0pt}{0pt}{1pt}

\usepackage{todonotes}

\usepackage{url}

\usepackage{algorithmic}
\usepackage{algorithm}
\usepackage{gensymb}
\usepackage{bm}

\newcounter{algsubstate}

\newlength\myindent
\begin{document}

\title{Quadrotor Neural Dead Reckoning in Periodic Trajectories}

\author{Shira~Massas~and~Itzik~Klein\\
        The Hatter Department of Marine Technologies, Charney School of Marine Sciences, University of Haifa.}


\maketitle

\begin{abstract}
In real world scenarios, due to environmental or hardware constraints, the quadrotor is forced to navigate in pure inertial navigation mode while operating indoors or outdoors. To mitigate inertial drift, end-to-end neural network approaches combined with quadrotor periodic trajectories were suggested. There, the quadrotor distance is regressed and combined with inertial model-based heading estimation, the quadrotor position vector is estimated. To further enhance positioning performance, in this paper we propose a quadrotor neural dead reckoning approach for quadrotors flying on periodic trajectories. In this case, the inertial readings are fed into a simple and efficient network to directly estimate the quadrotor position vector. Our approach was evaluated on two different quadrotors, one operating indoors while the other outdoors. Our approach improves the positioning accuracy of other deep-learning approaches, achieving an average 27\% reduction in error outdoors and an average 79\% reduction indoors, while requiring only software modifications. With the improved positioning accuracy achieved by our method, the quadrotor can seamlessly perform its tasks.
\end{abstract}

\section{Introduction}\label{intro_sec}
A quadrotor is an unmanned aerial vehicle with four rotors that provide lift, stability, and control for vertical takeoff, hovering, and maneuvering \cite{barcelos2020aerodynamic,rehan2022vertical}.
Quadrotors are used for a growing variety of applications, such as construction, transportation, industrial use, surveillance, law enforcement, emergency response, mapping, and more \cite{wang2024additive,bouras2023concentration}. To complete their tasks, accurate and robust navigation is critical.
While indoors,commonly, the inertial sensors are fused with cameras \cite{najafi2019adaptive,turci2024comparison} to obtain accurate navigation. One of the leading solutions for INS/camera fusion is simultaneous localization and mapping (SLAM) \cite{zheng2025visual,jiang2019simultaneous}. While outdoors the quadrotor navigation system relies on the fusion between the global navigation satellite systems (GNSS) and inertial sensors~\cite{ye2023review}.\\
Challenges in quadrotor navigation include poor visual conditions, distortion, or camera malfunctions while operating indoors \cite{mendoza2019meta} and and global navigation satellite systems (GNSS) outages while outdoors \cite{zidan2020gnss}. Additionally, GNSS-denied environments, such as tunnels or dense urban areas, pose significant challenges in obtaining accurate navigation \cite{atia2019map}.  In such scenarios, navigation often relies entirely on the inertial navigation system (INS) for positioning, in a situation known as pure inertial navigation \cite{9101092,titterton2004strapdown}. However, this approach navigation solution suffers from significant drift over time due to noise and errors in the inertial sensors. \\
One of the means to reduce the inertial drift in situations of pure inertial navigation, is to fly the quadrotor in periodic trajectories (PTs).  The quadrotor dead reckoning was the first model-based approach utilizing PTs~\cite{shurin2020qdr}. The QDR method evolved from pedestrian dead reckoning (PDR) methods where wearable inertial sensors are used to determine pedestrian position. While in PDR, periodic motion is a natural result of human movement, in QDR, PTSs are enforced to enable accurate navigation. In this manner, the peak-to-peak (period) distance of the quadrotor can be accurately estimated, similar to how step length is detected and estimated in PDR. With the advancement of deep learning techniques, the offer improved robustness, adaptability, and contribute in improving positioning accuracy \cite{chen2024deep,cohen2024inertial}.
QuadNet is a hybrid deep-learning framework for quadrotor dead reckoning that estimates three-dimensional position only by using inertial sensors~\cite{shurin2022quadnet}. Using regression neural networks, QuadNet calculates the quadrotor's change in distance and altitude, and determines its heading based on model-based equations. To further improve QuadNet performance, a multiple inertial sensor approach was suggested in~\cite{hurwitz2023quadrotor}. Later, in~\cite{hurwitz2024deep} a deep learning network, based on QuadNet, was proposed. With reduced number of layers and parameters their framework showed superior performance over QuadNet. In addition, they demonstrated that the regressed position vector can be used as a measurement update to the inertial navigation system in an extended Kalman filter fusion process. Recently, Aizelman et al. \cite{aizelman2024quadrotor} demonstrated the effectiveness of the QDR approach in terms of the trade-off between battery power consumption and navigation solution accuracy. However, all these methods dependent on estimated model-based heading angle , which introduces a limitation on performance. \\
To fill this gap, we propose QuadPosNet a novel quadrotor end-to-end neural dead reckoning approach that regresses the change in the quadrotor position using only its inertial sensors, and thus neglecting the need for the INS-based heading angle.
%
To evaluate our proposed approach, we adapt the outdoor dataset from \cite{hurwitz2023quadrotor} and use our newly recorded indoor dataset. In total we use $56.2$ minutes of inertial recordings from the two datasets. Our key result demonstrates that our proposed frameworks outperform the baseline (QuadNet) by 27\% on the outdoor dataset and by 79\% on the indoor dataset.
These results demonstrate the robustness and effectiveness of our proposed framework for pure inertial navigation. It is particularly useful for real-world applications in which GNSS signals are unreliable or not present at all. \\
The rest of the paper is structured as follows: In Section \ref{sec:problem}, we provide an overview of inertial navigation and the QuadNet approach. Section \ref{sec:prop} introduces our two proposed neural inertial framework. Section \ref{res_section} describes the datasets and gives the results. Finally, Section \ref{conc_sec} concludes this research.
\section{Problem Formulation}\label{sec:problem}
\subsection{Inertial Navigation}\label{INS_sec}
In pure inertial navigation, the INS mechanism provides the navigation solution, namely the position, velocity, and orientation \cite{titterton2004strapdown}. When the Earth's rotation rate is neglected, due to the use of low-cost, low-performance inertial sensors that are unable to detect the Earth's rotation, the equations of motion for the INS include the position vector rate of change, the velocity vector rate of change, and the transformation matrix between the body and navigation frame rate of change:
\begin{eqnarray}
\boldsymbol{\dot{p}}^{n} &=& \boldsymbol{X}^{n} \label{eq:INSEOM1}
    \label{eq:INSeq1}\\
\boldsymbol{\dot{v}}^{n} &=& \boldsymbol{T}^{n}_{b}\boldsymbol{f}_{ib}^{b} + \boldsymbol{g}^{n}\label{eq:INSEOM2}
    \label{eq:INSeq2}\\
\boldsymbol{\dot{T}}^{n}_{b} &=& \boldsymbol{T}^{n}_{b}\boldsymbol{\Omega}_{ib}^{b}\label{eq:INSEOM3}
    \label{eq:INSeq3}
\end{eqnarray}
where $\boldsymbol{p}^{n}$ is the position vector expressed in the local navigation frame, $\boldsymbol{X}^{n}$ is the velocity vector expressed in the navigation frame, $\boldsymbol{g}^{n}$  is the local gravity vector expressed in the navigation frame,  $\boldsymbol{\Omega}_{ib}^{b}$ is a skew-symmetric form of the angular velocity vector expressed in the body frame, $\boldsymbol{f}_{ib}^{b}$ is the specific force vector expressed in the body frame, and  $\boldsymbol{T}^{n}_{b}$ \eqref{eq:Teq} is the transformation matrix from body to navigation frame:\\
\begin{equation}\label{eq:Teq}
\boldsymbol{T}^{n}_{b}=\begin{bmatrix}C_{\theta}C_{\psi}&S_{\phi}S_{\theta}C_{\psi}-C_{\phi}S_{\psi}&C_{\phi}S_{\theta}C_{\psi}+S_{\phi}S_{\psi}\\C_{\theta}S_{\psi}&S_{\phi}S_{\theta}S_{\psi}+C_{\theta}C_{\psi}&C_{\phi}S_{\theta}S_{\psi}-S_{\phi}C_{\psi}\\-S_{\theta}&S_{\phi}C_{\theta}&C_{\phi}C_{\theta}\end{bmatrix}
\end{equation}\\
where $\boldsymbol{T}^{n}_{b}[i,j]$ is the element in the i row and j column, $S_{x}$ is the sine of $x$, and $C_{x}$ is the cosine of $x$, and $\phi$, $\theta$ and $\psi$ are the roll, pitch and yaw angles respectively, and the $\psi$ angle \cite{9101092} is:


\begin{equation}\label{eq:psi1}
\psi = \arctan2(\boldsymbol{T}^{n}_{b}[1,2], \boldsymbol{T}^{n}_{b}[1,1])
\end{equation}

\subsection{The QuadNet Method}\label{QDR_subsec}
The QuadNet model is a regression network capable of estimating the quadrotor's distance and altitude \cite{shurin2022quadnet}. The QuadNet architecture consists of five 1D-convolution layers leading to the fully connected layers.

It leverages neural network structures to learn and extract features from the sensor data, enabling accurate regression of the quadrotor's position parameters.
To calculate the next quadrotor's position using the QuadNet method, the predicted regression of the quadrotor's position parameters is utilized. To calculate the quadrotor's horizontal positioning we use:
\begin{eqnarray}\label{eq:qdrpos}
	\textit{x}_{k+1} & = & \textit{x}_{k}+\textit{d}_{k}\cos\psi_{k}\\
	\textit{y}_{k+1} & = & \textit{y}_{k}+\textit{d}_{k}\sin\psi_{k}\label{eq:qdrpos2}
\end{eqnarray}
where $\textit{x}_k$ and $\textit{y}_k$ are the current position components in the $x$ and $y$ directions, respectively, and the $\psi_k$ is the calculated INS heading angle \eqref{eq:psi1}, where \eqref{eq:qdrpos} is used to determine the \(x\)-component of the position, while \eqref{eq:qdrpos2} defines the \(y\)-component.

\section{Proposed Approach}\label{sec:prop}
\subsection{Architecture}\label{Architecture_sec}
\begin{figure}[H]
	\centering
	\includegraphics[width=1\linewidth]{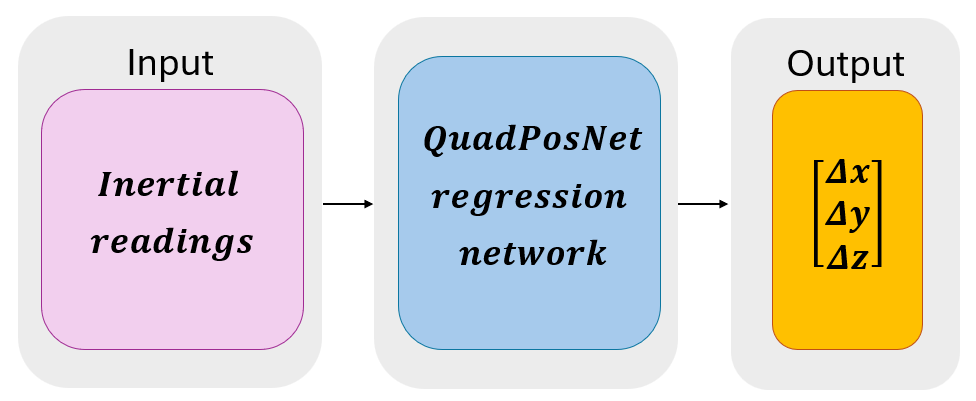}
    \caption{Fig. Our QuadPosNet end-to-end neural network architecture. The network processes inertial measurements (accelerometer and gyroscope raw data) and predicts the change in the quadrotor's position vector.}
	\label{fig:scheme11}	
\end{figure}

In this research, we propose a method to reduce position drift in quadrotors relying solely on inertial navigation. We introduce QuadPosNet, a neural dead reckoning regression network that takes raw inertial measurements and outputs changes in the quadrotor's position, as shown in Figure~\ref{fig:scheme11}. We examine two network architectures for the regression task: a single-head QuadPosNet network that combines accelerometer and gyroscope data as a single input, and a multi-head QuadPosNet network that processes each data separately through two input heads. Both networks employ one-dimensional convolutional and fully connected layers, with the change in the quadrotor position vector as the output.\\
Our QuadPosNet frameworks, like QuadNet, predict changes in a quadrotor's position based on inertial data. However, unlike QuadNet, which predicts a two-dimensional position and requires additional calculations involving a computed angle, our frameworks directly predict the position vector along all three axes simultaneously. This approach eliminates the need for further computations, thereby reducing the potential for additional errors in position prediction.

\subsubsection{Single-Head QuadPosNet}\label{single_subsec}

The input to the single-head QuadPosNet network is a multi-dimensional array representing a time series of raw inertial measurement units (IMU) data from the quadrotor, consisting of six channels: three accelerometer readings and three gyroscope readings. The network combines these six channels into a single input to predict the change in the quadrotor's position vector.

\begin{figure}[H]
	\centering
	\includegraphics[width=1\linewidth]{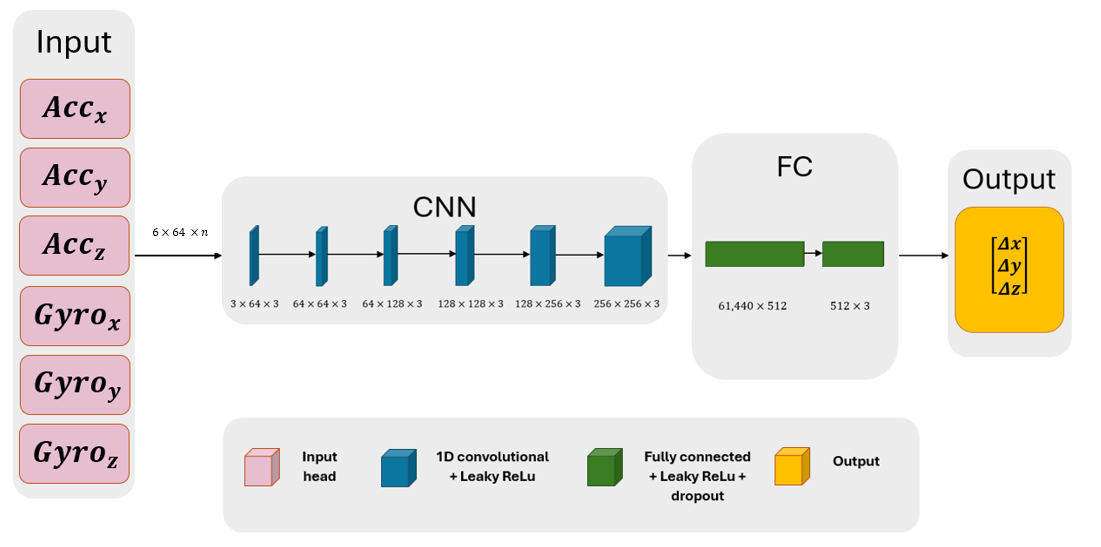}
	\caption{Fig. Single-head QuadPosNet architecture consisting of 1D-convolutional layers, used for feature extraction, and fully connected layers that output of quadrotor change in the position vector.}
	\label{fig:BD1}	
\end{figure}

The single-head network architecture is shown in Figure~\ref{fig:BD1}. The pink layer is the input head, each input is structured as a three-dimensional tensor of shape \( 64 \times 6 \times n \), where $64$ represents the batch size, $6$ corresponds to the number of sensor channels, and $n$ is the window size. the input is then passed through six one-dimensional convolutional layers with leaky ReLU activation functions.

Following the convolutional layers, the extracted features are passed through two fully connected layers with leaky ReLU activation functions and a dropout regularization. Finally, the output layer estimates the change in the quadrotor's position vector.  

The input to the single-head network is a one-dimensional array (time series) representing raw inertial measurements is:

\begin{equation}
\mathbf{X} = \{f_x, f_y, f_z, \omega_x, \omega_y, \omega_z\} \in \mathbb{R}^{6 \times n}
\label{eq:input_array}
\end{equation}

where \(f_x, f_y, f_z\) are the accelerometer readings along the \(x\), \(y\), and \(z\) axes, respectively, and \(\omega_x, \omega_y, \omega_z\) are the gyroscope readings along the \(x\), \(y\), and \(z\) axes, respectively.

The activation function employed, both in the convolutional layers and in the fully connected layers, is the leaky rectified linear unit (Leaky ReLU). It is an activation function that allows a small, non-zero gradient when the input is negative, while preserving positive values unchanged \cite{dubey2022activation}:

\begin{equation}
\text{LeakyReLU}(x) = 
\begin{cases} 
x, & \text{if } x \geq 0 \\
\alpha \cdot x, & \text{otherwise}
\end{cases}
\label{eq:leaky_relu}
\end{equation}

where \( \alpha \) is a small positive constant, used as a scaling factor for negative values. A typical value for \( \alpha \) is \( 0.01 \). This function helps avoid the vanishing gradient problem that often occurs in deep neural networks by allowing a small gradient to flow when the input is negative.

The hidden layers of the neural network are defined recursively. The first hidden layer, \( h_1 \) \eqref{eq:hidden_layer1} is computed by using a Leaky ReLU activation function \cite{liu2023inexact} applied to the linear transformation of the input \( \mathbf{X} \):

\begin{equation}
h_1 = \text{LeakyReLU} \left( \mathbf{W}_0 \mathbf{X} + \mathbf{b}_0 \right)
\label{eq:hidden_layer1}
\end{equation}

where \( \mathbf{W}_0 \) is the initial weight matrix and \( \mathbf{b}_0 \) is the initial bias vector. The rest of the hidden layers \( h_n \) for \( n = 2, \dots, N \) are:

\begin{equation}
\begin{split}
h_n &= \text{LeakyReLU} \left( \mathbf{W}_{n-1} h_{n-1} + \boldsymbol{b}_{n-1} \right), \\
&\hfill n = 2, \dots, N \hfill
\end{split}
\label{eq:hidden_layer_n}
\end{equation}

where \( \mathbf{W}_{n-1} \) and \( \mathbf{b}_{n-1} \) are the weights and biases of the \( (n-1) \)-th layer.

The convolutional layers in the single-head network process the input recursively:

\begin{equation}
h_{6,\text{out}} = h_6 \left( h_5 \left( h_4 \left( h_3 \left( h_2 \left( h_1 \right) \right) \right) \right) \right)
\label{eq:cnn_layers}
\end{equation}

Each convolutional layer \( h_2, h_3, \dots, h_6 \) successively transforms the features extracted by the previous layer, enabling the network to capture hierarchical representations of the input data, and \(h_{6,\text{out}}\) represents the output from the consecutive convolutional layers.

The output of the network, $\Delta \textbf{p}$,  is the change in the quadrotor position and is obtained after passing two fully connected layers:

\begin{equation}
\Delta \textbf{p} = h_8 \left( h_7 \left( h_{6,\text{out}} \right) \right)
\label{eq:single_head_output}
\end{equation}

where \( h_7 \) and \( h_8 \) are two fully connected layers, and $\Delta \textbf{p}$ is the predicted change in position of the quadrotor defined by:

\begin{equation}
\Delta \mathbf{p} = 
\begin{bmatrix} 
\Delta x \\ 
\Delta y \\ 
\Delta z 
\end{bmatrix}
\label{eq:final_output}
\end{equation}

where \( \Delta x \) is the change in position along the \( x \)-axis, \( \Delta y \) is the change in position along the \( y \)-axis, and \( \Delta z \) is the change in position along the \( z \)-axis.

\subsubsection{Multi-Head QuadPosNet}\label{multi_subsec}

The input to the multi-head QuadPosNet network consists of two separate multi-dimensional arrays representing time series of raw inertial data from the quadrotor: one array contains three accelerometer readings, while the other holds three gyroscope readings. Unlike the single-head model, which merges all six channels into one input, the multi-head network processes each sensor's data separately, leveraging accelerometer and gyroscope features to predict changes in the quadrotor's position vector.

\begin{figure}[H]
	\centering
	\includegraphics[width=1\linewidth]{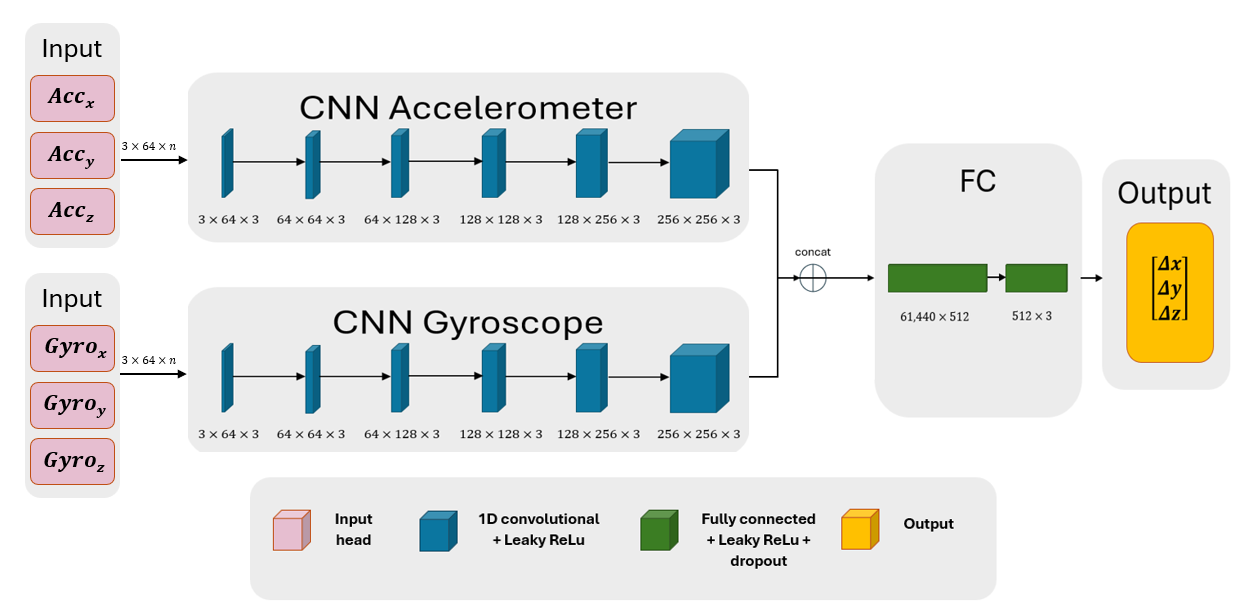}
	\caption{Fig. Multi-head QuadPosNet architecture consisting has two input heads, each consists of 1D-convolutional layers, used for feature extraction, and fully connected layers that output of quadrotor change in the position vector.}
	\label{fig:BD2}	
\end{figure}

The multi-head QuadPosNet network architecture is shown in Figure~\ref{fig:BD2}. The pink layers represent the input heads for the accelerometer and gyroscope data. Each input is structured as a three-dimensional tensor with the shape \( 64 \times 3 \times n \), where \( 64 \) represents the batch size, \( 3 \) corresponds to the number of sensor channels, and \( n \) is the window size. Each branch processes its input through six one-dimensional convolutional layers, each equipped with leaky ReLU activation functions. These layers independently extract features specific to the accelerometer and gyroscope data.

After the sixth convolutional layer, the outputs of both branches are concatenated into a unified representation. This combined representation is then passed through a fully connected layer with leaky ReLU activation and a dropout regularization. Another fully connected layer with leaky ReLU activation and dropout follows. Finally, the output layer predicts the quadrotor’s change in position vector.

The input to the multi-head QuadPosNet network consists of two separate one-dimensional arrays (time series) representing raw inertial measurements:

\begin{equation}
\mathbf{X}_{\text{acc}} = \{f_x, f_y, f_z\} \in \mathbb{R}^{3 \times n}
\label{eq:acc_input}
\end{equation}

\begin{equation}
\mathbf{X}_{\text{gyro}} = \{\omega_x, \omega_y, \omega_z\} \in \mathbb{R}^{3 \times n}
\label{eq:gyro_input}
\end{equation}

The two inputs, \( \mathbf{X}_{\text{acc}} \) and \( \mathbf{X}_{\text{gyro}} \), are processed independently through separate convolutional layers before being combined and processed through fully connected layers.

The activation function employed, both in the convolutional layers and the fully connected layers, is the Leaky ReLU \eqref{eq:leaky_relu}.

The convolutional layers for both the accelerometer and gyroscope heads process their respective inputs recursively, as defined for the single-head case \eqref{eq:cnn_layers}. Specifically, the input \( \mathbf{X}_{\text{acc}} \) for the accelerometer head and the input \( \mathbf{X}_{\text{gyro}} \) for the gyroscope head are independently processed through six one-dimensional convolutional layers with Leaky ReLU activations. Each layer transforms the features extracted by the previous layer, enabling the network to capture hierarchical representations of the accelerometer and gyroscope data. 

After the sixth convolutional layer, the outputs of the accelerometer and gyroscope separated last convolutional layers are concatenated to combine their feature representations:

\begin{equation}
h_{\text{concat}} = \text{concat} \left( h_{\text{acc},6}, h_{\text{gyro},6} \right)
\label{eq:concat_layer}
\end{equation}

Each convolutional layer \( h_2, h_3, \dots, h_6 \) successively transforms the features extracted by the previous layer, enabling the network to capture hierarchical representations of the input data.  
The unified feature representation \( h_{\text{concat}} \) integrates information from both the accelerometer and gyroscope, thus creating a unified feature representation for the next layers.

The output of the network, $\Delta \bm{p}$  is the change in the quadrotor position and is obtained after passing two fully connected layers:

\begin{equation}
\Delta \bm{p} = h_8 \left( h_7 \left( h_{\text{concat}} \right) \right)
\label{eq:multi_head_output}
\end{equation}

where \( h_7 \) and \( h_8 \) are two fully connected layers, and \( h_{\text{concat}} \) represents the unified feature representation obtained by combining the outputs of the accelerometer and gyroscope convolutional layers, and $\Delta \textbf{p}$ is the predicted change in position \eqref{eq:final_output}. 

\subsection{Training Process}\label{train_sec}

\subsubsection{Adam Optimization Algorithm}\label{adam_opt_sec}
We employ the adaptive moment estimation (Adam) \cite{diederik2014adam}. It is a method that computes adaptive learning rates for each parameter. 
The exponentially decaying averages of past gradients 
\begin{equation}
\mathbf{m}_t = \beta_1 \mathbf{m}_{t-1} + (1 - \beta_1) \mathbf{g}_t,
\label{eq:adam_mt}
\end{equation}
and squared gradients 
\begin{equation}
\mathbf{v}_t = \beta_2 \mathbf{v}_{t-1} + (1 - \beta_2) \mathbf{g}_t^2
\label{eq:adam_vt}
\end{equation}
Here, \( \mathbf{m}_t \) and \( \mathbf{v}_t \) are estimates of the first moment  and the second moment of the gradients, respectively. As \( \mathbf{m}_t \) and \( \mathbf{v}_t \) are first initialized as vectors of zeros, they can become biased towards zero, especially during the initial time steps when the decay rates \( \beta_1 \) and \( \beta_2 \) are close to 1. To counteract biases, the bias-corrected moment estimates of \( \mathbf{m}_t \) and of \( \mathbf{v}_t \)  are:

\begin{equation}
\hat{\mathbf{m}}_t = \frac{\mathbf{m}_t}{1 - \beta_1^t}
\label{eq:adam_mt_corrected}
\end{equation}
\begin{equation}
\hat{\mathbf{v}}_t = \frac{\mathbf{v}_t}{1 - \beta_2^t}
\label{eq:adam_vt_corrected}
\end{equation}

These estimates are then used to update the parameters according to the Adam update rule:

\begin{equation}
\boldsymbol{\theta}_{t+1} = \boldsymbol{\theta}_t - \frac{\eta}{\sqrt{\hat{\mathbf{v}}_t} + \epsilon} \hat{\mathbf{m}}_t
\label{eq:adam_update}
\end{equation}

where \( \boldsymbol{\theta} \) represents the parameters of the model that are being optimized, \( \eta \) is the learning rate, which controls the step size of the parameter updates, and \( \epsilon \) is a small constant added to the denominator to ensure numerical stability and prevent division by zero.

\subsubsection{Loss Function}\label{loss_function_sec}

The mean squared error (MSE) is used as the loss function, which is particularly suitable for regression tasks where continuous values are predicted. The MSE loss is defined by:

\begin{equation}
J(\mathbf{o}_i, \hat{\mathbf{o}}_i) = \frac{1}{n} \sum_{i=1}^{N} \| \mathbf{o}_i - \hat{\mathbf{o}}_i \|^2
\label{eq:mse}
\end{equation}

where \( J() \) represents the MSE function, \( \mathbf{o}_i \) is the ground truth (GT) vector, \( \hat{\mathbf{o}}_i \) is the network’s estimated vector, and \( n \) denotes the number of samples.
The loss function minimizes the squared differences between predicted and true values, ensuring robustness for varied trajectories.

\section{Experiment Setup and Results}\label{res_section}
\subsection{Dataset}\label{dataset_subsec}
Two datasets recorded with different quadrotors in different environments are used for our analysis. The first one is an outdoor dataset recorded using DJI Phantom 4 GNSS RTK quadrotor \cite{hurwitz2023quadrotor}. We recorded the second dataset indoors using the Crazyflie 2.1 quadcopter \cite{crazyflie2024}.

\subsubsection{Outdoor Dataset}\label{gnss_rtk_subsec}
\begin{figure}[H]
	\centering
	\includegraphics[width=1\linewidth]{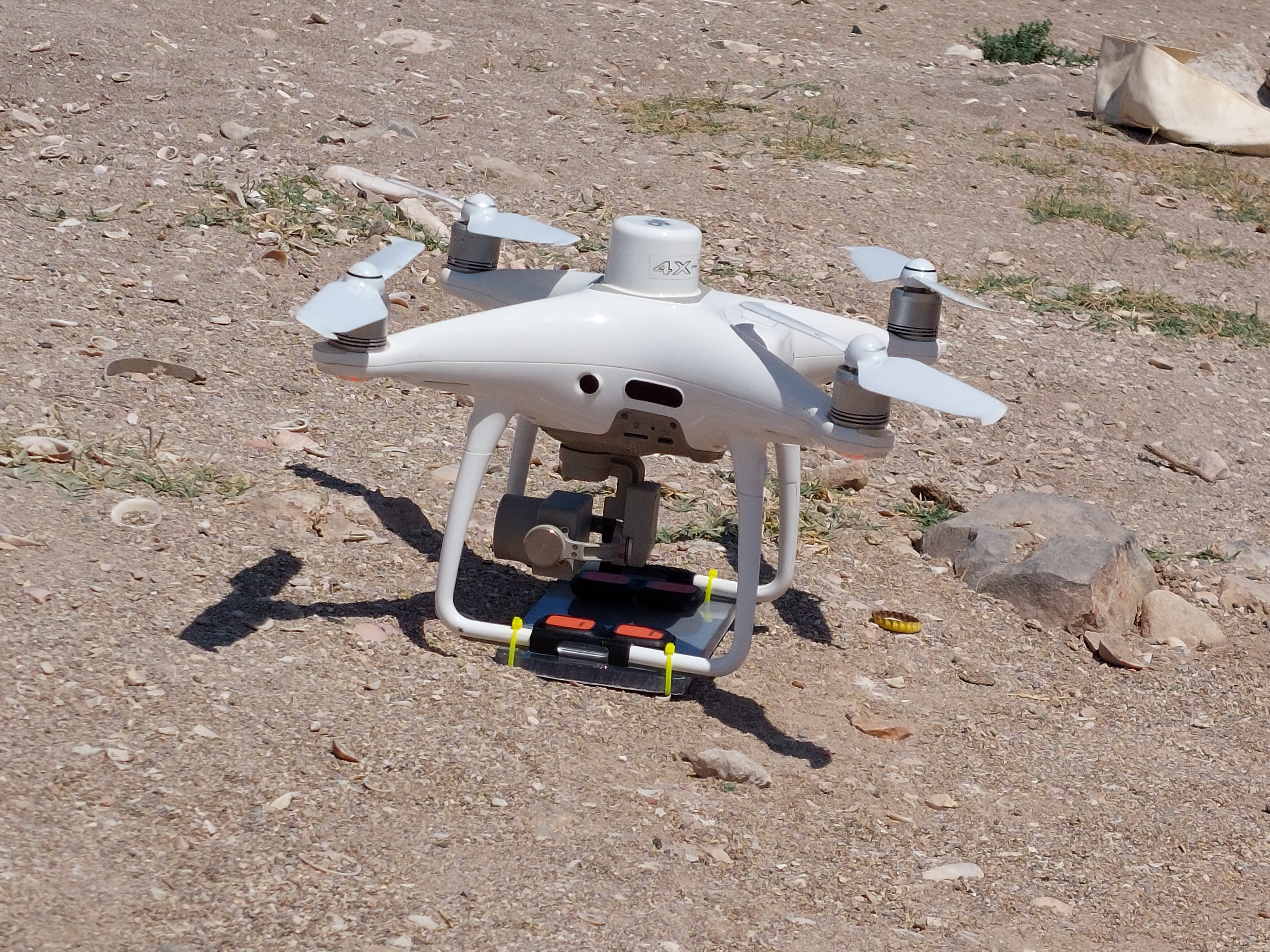}
    \caption{Fig. Outdoor experimental setup featuring a DJI Phantom 4 quadrotor equipped with four IMUs in our field experiment setup.}
	\label{fig:drone2}	
\end{figure}

The outdoor dataset includes vertical (D1, D2) and horizontal (D3) trajectories recorded with a DJI Phantom 4 GNSS RTK quadrotor, equipped with four Dot sensors \cite{movellaDot} as shown in Figure~\ref{fig:drone2}, from which one was used in our data. The GT data was gathered from the quadrotor itself (RTK positioning), while the IMUs served as the units under test. These experiments were made outdoors with the quadrotor operated by a human pilot. These trajectories were selected to represent diverse flight conditions and ensure robustness in GNSS-denied scenarios.

The Dot sensor is a compact and lightweight device that can be easily configured via Bluetooth Low Energy (BLE) 5.0. It utilizes the BOSCH BNO055 sensor module. In our researched we use the measurement of one IMU on it's platform. In total the dataset contains of $36.4$ minutes of inertial data sampled at $120$[Hz] and associated position GT.\\

The dataset is divided into:
\begin{itemize}
\item \textbf{D1 - Vertical 1:} Eight trajectories with a combined duration of $5.8$ minutes.
\item \textbf{D2 - Vertical 2:} Thirty trajectories lasting a total of $16.2$ minutes.
\item \textbf{D3 - Horizontal:} Twenty-seven trajectories spanning a total time of $14.4$ minutes.
\end{itemize}
The vertical recordings are divided into D1 and D2 as they were recorded in different days.

\subsubsection{Indoors Dataset}\label{crazy_subsec}
\begin{figure}[H]
    \centering
    \includegraphics[width=1\linewidth]{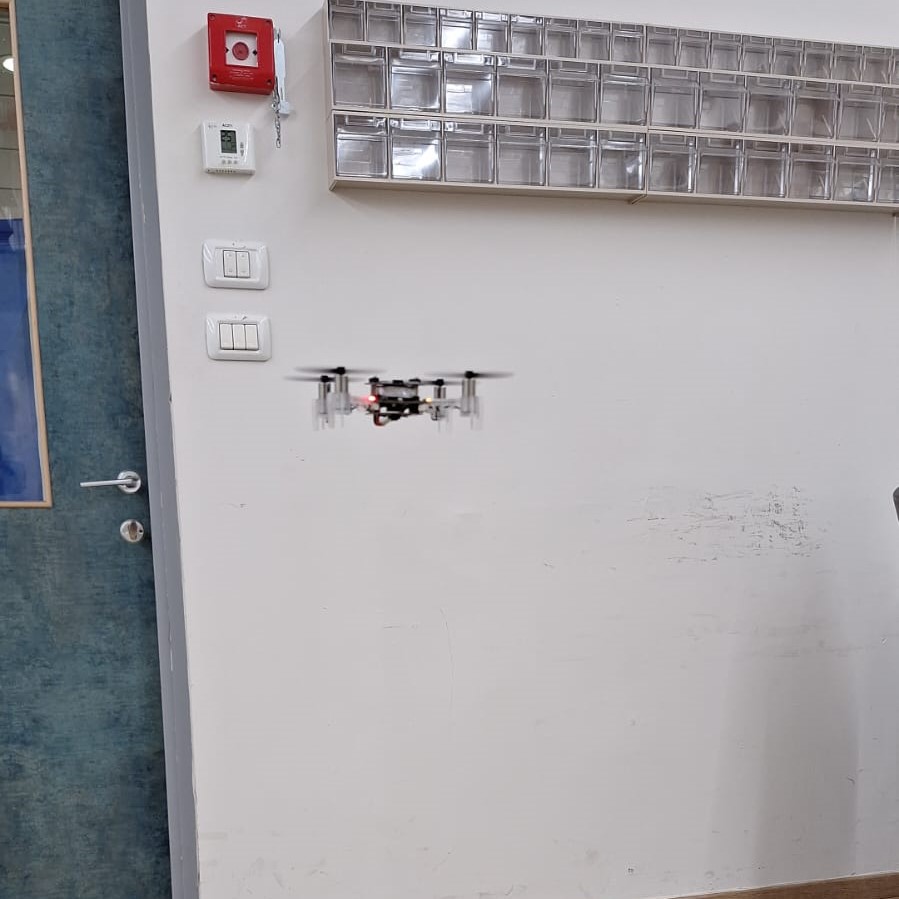}
    \caption{Fig. Indoor experimental setup showing the Crazyflie quadrotor during data collection.}
    \label{fig:crazy3}    
\end{figure}

For the creation of our indoor dataset we used the Crazyfile 2.1 quadcopter. The Crazyflie weighs $27$[g], it's battery life is $7$ minutes, it's charging time with stock battery is $40$ minutes and it's measurements are \( 92 \times 92 \times 29 \, \text{mm} \), and is it captured in Figure~\ref{fig:crazy3}. The Crazyflie has the BMI088 IMU model.
GT position measurements are obtained using the lighthouse positioning system with two SteamVR $2.0$ virtual reality stations. The GT position accuracy is $2$-$4$[cm].

\begin{figure}[H]
	\centering
	\includegraphics[width=1\linewidth]{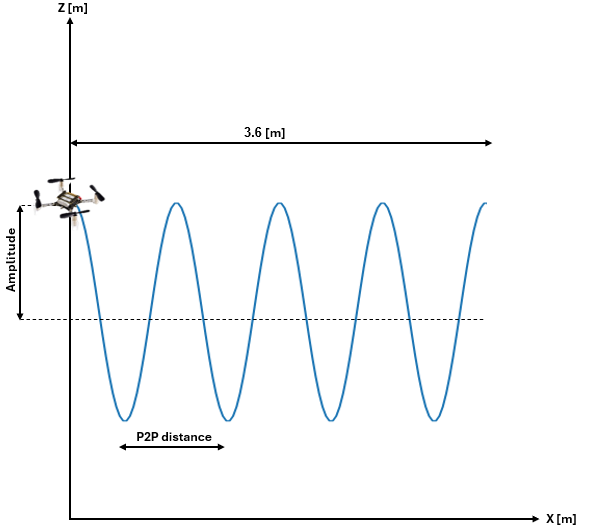}
\caption{Fig. An illustration of the sinusoidal flight trajectory of the Crazyflie. The figure shows the amplitude and peak-to-peak (P2P) distance, key parameters for evaluating position estimation performance.}
    \label{fig:crazy2}	
\end{figure}

The Crazyflie’s trajectory in our datasets followed a sinusoidal pattern, as shown in Figure~\ref{fig:crazy2}. At Dataset D4 the Crazyflie started by hovering at $0.5$[m], the trajectory had an amplitude of $0.65$[m] and a peak-to-peak (P2P) distance of $0.9$[m], while at Dataset D5 the Crazyflie started by hovering at $0.7$[m], featured an amplitude of $0.1$[m] and a P2P of $0.7$[m]. These parameters were deliberately varied to simulate diverse flight conditions and make two distinctive datasets. Both experiments started from the same fixed point and covered a consistent distance of $3.6$[m] along the x-axis. In total the dataset contains of $19.4$ minutes of inertial data sampled at 100[Hz] and associated position GT.

\begin{itemize}
\item \textbf{D4 - Crazyflie Vertical 1:}
$45$ trajectories with a combined duration of $14.2$ minutes.
\item \textbf{D5 - Crazyflie Vertical 2:}
$35$ trajectories with a combined duration of $5.2$ minutes.
\end{itemize}

By adjusting the amplitude and P2P values, we ensured that the dataset captured a variety of flight conditions, providing a robust foundation for analysis and model validation. Both experiments were conducted indoors, with the Crazyflie drone starting from the same fixed point each time and covering a consistent distance of $3.6$[m] along the $x$-axis.

\subsubsection{Summary}\label{summerydata_subsec}

Table~\ref{tab:datasets_table} presents the division of our dataset. 
For the outdoor dataset, the training used a window size of $120$ and a stride of $60$, sampled at $120$[Hz]. Indoor datasets used a window size of $100$ and a stride of $50$, sampled at $100$[Hz]. For the outdoor datasets, the training dataset for D1 consists of six trajectories, while the testing dataset contains two trajectories that were not used in training. For D2 and D3, the training dataset includes $26$ and $23$ trajectories, respectively, while the testing dataset contains four trajectories that were excluded from the training data. Similarly, for the indoor datasets, the training dataset for D4 and D5 include of $41$ and $31$ trajectories, respectively, while the testing datasets contains four trajectories that were excluded from the training data.

\begin{table}[H]
\centering
\caption{Fig. Training and testing times for indoor and outdoor datasets, including stride and window size.}
\renewcommand{\arraystretch}{1.2}
\resizebox{0.5\textwidth}{!}{
\begin{tabular}{|c|c|c|c|c|c|}
\hline
\textbf{Dataset} & \textbf{Train Set} & \textbf{Test Set} & \textbf{Total} & \textbf{Stride} & \textbf{Window Size} \\ 
\hline
D1 Vertical 1 (Outdoor) & 4.36 & 1.44 & 5.8 & 60 & 120 \\ \hline
D2 Vertical 2 (Outdoor) & 15.1 & 1.1 & 16.2 & 60 & 120 \\ \hline
D3 Horizontal (Outdoor) & 13.3 & 1.1 & 14.4 & 60 & 120 \\ \hline
D4 Vertical 1 (Indoor) & 13.6 & 0.6 & 14.2 & 50 & 100 \\ \hline
D5 Vertical 2 (Indoor) & 4.9 & 0.3 & 5.2 & 50 & 100 \\ \hline
\textbf{Total} & \textbf{51.26} & \textbf{4.54} & \textbf{55.8} & -- & -- \\ 
\hline
\end{tabular}
}
\label{tab:datasets_table}
\end{table}

\subsection{Evaluation Metric}\label{eval_subsec}

The root mean squared error (RMSE) metric was used for comparison between the different approaches.  
A lower RMSE value indicates better model performance, reflecting less variance between expected and actual values. The RMSE is defined by:

\begin{equation}
\text{RMSE} = \sqrt{\frac{1}{n} \sum_{i=1}^{n} (\mathbf{p}_i - \hat{\mathbf{p}}_i)^2}
\label{eq:rmse}
\end{equation}

where \( \mathbf{p}_i \) is the GT position vector, \( \hat{\mathbf{p}}_i \) is the predicted position, and \( n \) is the number of samples.

\subsection{Experimental Results}\label{hyper_subsec}

In the outdoor and the indoor datasets and for both network architectures a batch size of $64$ and learning rate of $1$e-$3$ were used. The training and testing procedures were repeated three times. The next tables shows the average of those three runs.

\subsubsection{\textbf{Outdoor Results}}\label{EXPERIMENT_outdoors_results}
Table~\ref{tab:comparison1} presents the performance results of our proposed methods, the single-head QuadPosNet and the multi-head QuadPosNet, on three outdoor datasets (D1-D3).  The table also shows the improvement achieved by our approaches compared to the QuadNet baseline.
Our findings clearly demonstrate the superiority of our proposed frameworks. On average, the single-head QuadPosNet improved upon the QuadNet baseline by $18.0\%$.  Even more impressively, the multi-head QuadPosNet achieved a $27.1\%$ average improvement over the baseline, outperforming all other tested methods. The substantial $27.1\%$ reduction in RMSE observed for the multi-head QuadPosNet across datasets D1-D3 highlights its effectiveness in mitigating drift errors.  This is particularly significant in outdoor environments where GNSS signals are unavailable.  This result shows the framework's capability to maintain accurate positioning for short durations without relying on external corrections, a crucial characteristic for robust autonomous navigation in challenging scenarios.

\begin{table}[ht]
\centering
\caption{Fig. Comparison of RMSE and improvement for different methods on the outdoor datasets.}
\renewcommand{\arraystretch}{1.2}
\resizebox{0.5\textwidth}{!}{
\begin{tabular}{|l|l|c|c|}
\hline
\textbf{Dataset} & \textbf{Method} & \textbf{RMSE Avg. [m]} & \textbf{Improvement [\%]} \\ 
\hline
\multirow{3}{*}{D1 Vertical 1 (Outdoor)} & QuadNet (baseline) & 15.1 & -- \\ 
& MH-QuadPosNet (ours) & 13.8 & 8.6 \\ 
& SH-QuadPosNet (ours) & 19.5 & X \\ \hline
\multirow{3}{*}{D2 Vertical 2 (Outdoor)} & QuadNet (baseline) & 22.1 & -- \\ 
& MH-QuadPosNet (ours) & 17.6 & 20.4 \\ 
& SH-QuadPosNet (ours) & 21.6 & 2.3 \\ \hline
\multirow{3}{*}{D3 Horizontal (Outdoor)} & QuadNet (baseline) & 28.9 & -- \\ 
& MH-QuadPosNet (ours) & 13.8 & 52.2 \\ 
& SH-QuadPosNet (ours) & 14.0 & 51.6 \\ 
\hline
\end{tabular}
}
\label{tab:comparison1}
\end{table}

\begin{figure}[H]
	\centering
	\includegraphics[width=1\linewidth]{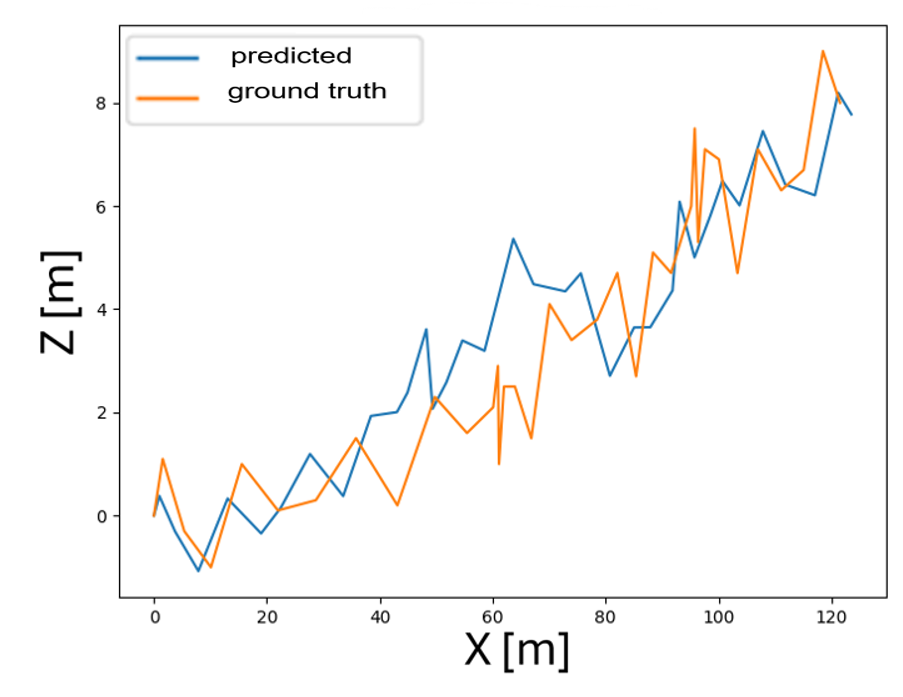}
	\caption{Fig. Vertical plane of a sample trajectory out of the D2 dataset showing the GT and the predicted trajectories. The latter was obtained with our multi-head architecture.}
    \label{fig:ver1}	
\end{figure}

Figure~\ref{fig:ver1} illustrates an example of a trajectory from the D2 dataset, showing the x vs. z position components. The y-component was omitted due to negligible variation. The predicted trajectory was generated using our multi-head architecture.

\subsubsection{\textbf{Indoor Results}}\label{EXPERIMENT}

Table \ref{tab:combined_dataset_comparison} provides the indoor datasets D4-D5 results as well as the improvement rate of our approaches relative to the baseline. The single-head QuadPosNet achieved a $79.4\%$ improvement, while the multi-head QuadPosNet achieved a $79.0\%$ improvement on average over the baseline. These results clearly demonstrate the effectiveness of our proposed frameworks in significantly outperforming the QuadNet method across both datasets. The $79.0\%$ improvement in RMSE for both single-head QuadPosNet and multi-head QuadPosNet on the indoor datasets D4 and D5 shows the framework's ability to handle precise trajectory prediction in controlled but challenging environments such as in poor visual conditions.

\begin{table}[ht]
\centering
\caption{Fig. Comparison of RMSE and improvement for different methods on the indoor datasets.}
\renewcommand{\arraystretch}{1.2}
\resizebox{0.5\textwidth}{!}{
\begin{tabular}{|l|l|c|c|}
\hline
\textbf{Dataset} & \textbf{Method} & \textbf{RMSE Avg. [m]} & \textbf{Improvement [\%]} \\ 
\hline
\multirow{3}{*}{D4 Vertical 1 (Indoor)} & QuadNet (baseline) & 1.84 & -- \\ 
& SH-QuadPosNet (ours) & 0.52 & 71.7 \\ 
& MH-QuadPosNet (ours) & 0.56 & 69.6 \\ \hline
\multirow{3}{*}{D5 Vertical 2 (Indoor)} & QuadNet (baseline) & 1.38 & -- \\ 
& SH-QuadPosNet (ours) & 0.18 & 87.0 \\ 
& MH-QuadPosNet (ours) & 0.16 & 88.4 \\ 
\hline
\end{tabular}
}
\label{tab:combined_dataset_comparison}
\end{table}

\begin{figure}[H]
	\centering
	\includegraphics[width=1\linewidth]{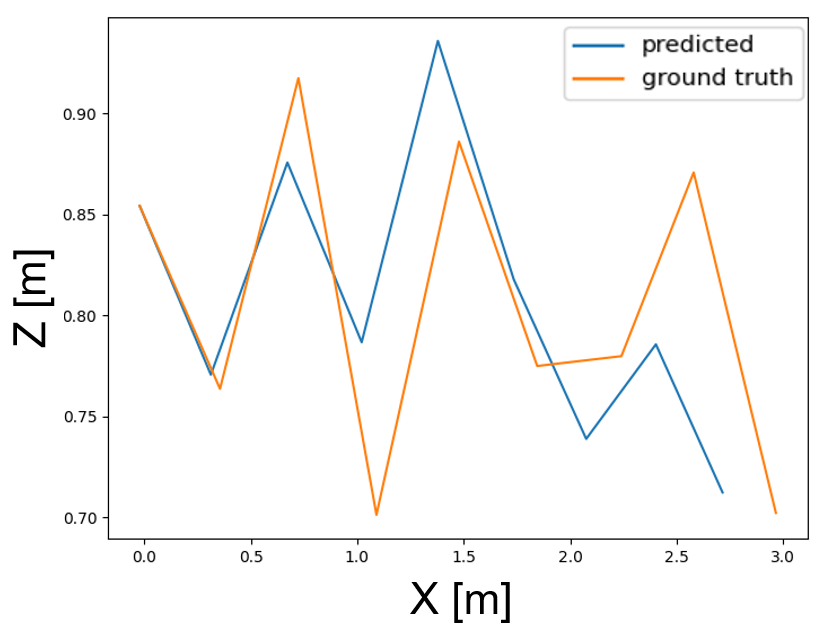}
	\caption{Fig. Vertical plane of a sample trajectory out of the D5 dataset showing the GT and the predicted trajectories. The latter was obtained with our multi-head architecture.}
    \label{fig:bbbbb}	
\end{figure}

Figure~\ref{fig:bbbbb} illustrates an example of a trajectory from the D5
dataset, showing the x vs. z components. The y-component
was omitted due to negligible variation. The predicted trajectory
was generated using our multi-head network architecture.

\section{Conclusion}\label{conc_sec}
In this research, we introduced two novel neural inertial network architectures, QuadPosNet (single-head) and QuadPos-Net (multi-head), designed to estimate changes in quadrotor position using only inertial readings.  This approach offers a significant advantage over standard methods in situations of pure inertial navigation when reliance on external positioning systems like GNSS or cameras is unavailable.  We rigorously evaluated the performance of our proposed architectures against the existing QuadNet baseline using two datasets, totaling 56.2 minutes of flight data.  One of these datasets was collected using a Crazyflie quadrotor in an indoor setting, providing a controlled environment for testing.

Our findings demonstrate substantial performance gains compared to the QuadNet baseline.  Specifically, in outdoor scenarios, the single-head QuadPosNet achieved an $18\%$ improvement in positional accuracy, while the multi-head QuadPosNet exhibited an even more impressive $27\%$ increase.  The benefits of our approach were particularly pronounced in indoor environments, where both the single-head and multi-head QuadPosNet architectures yielded a remarkable $79\%$ accuracy improvement.  These results underscore the robustness and effectiveness of our proposed frameworks for pure inertial navigation.

These advancements are particularly crucial for real-world applications where GNSS signals are unreliable or completely absent.  Examples include disaster response scenarios, urban delivery operations in dense cityscapes, and indoor navigation. By significantly enhancing the reliability of pure inertial navigation, our approach paves the way for more effective and autonomous deployment of quadrotors in challenging and GNSS-denied environments.

It is important to acknowledge a limitation of our current evaluation. Our focus was primarily on periodic motion trajectories, which were porven to be critical for improving the navigation solution in pure inertial navigation scenarios. While this strategy enhances accuracy, it introduces a trade-off: the increased processing and control activity associated with periodic motion can lead to higher battery consumption.  Therefore, future work will need to address the balance between accuracy and battery life to maximize mission duration and practical applicability in real-world deployments. Investigating methods for optimizing power consumption without sacrificing accuracy will be a key direction for future research.

\section{Aknowlegments}\label{conc_sec}
The authors are grateful to the support of Itai Savin with the Crazyflie experiments. S. M. was supported by the Maurice Hatter Foundation.

\bibliographystyle{IEEEtran}

\end{document}